\documentclass[10pt]{article} 
\usepackage[preprint]{tmlr}  
\usepackage[hyphens]{url}  
\usepackage{hyperref}
\usepackage{graphicx} 
\usepackage{natbib}  
\usepackage{caption} 
\usepackage{booktabs}
\usepackage{amsfonts, amsmath, amssymb, amsthm}
\usepackage{bbold}
\usepackage{algpseudocode, algorithm}
\usepackage{url, xcolor}
\usepackage{tikz}
\usetikzlibrary{fit}
\usetikzlibrary{arrows.meta, positioning, calc, shapes.geometric, shadows, backgrounds, decorations.pathreplacing}
\usepackage{subcaption}

\DeclareMathOperator{\vect}{vec}

\DeclareMathOperator*{\argmax}{arg\,max}

\DeclareMathOperator{\Span}{span}

\newtheorem{deff}{Definition}
\newtheorem{lem}{Lemma}
\newtheorem{thm}{Theorem}
\newtheorem{prop}{Proposition}

\newtheorem{cor}{Corollary}

\newcommand{\margarida}[1]{\textcolor{teal}{#1}}

\title{Bridging Neural Tangent Kernel Analysis and Reinforcement Learning for Deep Function Approximation}
\author{
    \name William St-Arnaud \email william.st-arnaud@umontreal.ca \\
    \addr Mila, Universit\'e de Montr\'eal
    \AND
    \name Margarida Carvalho \email carvalho@iro.umontreal.ca \\
    \addr Mila, Universit\'e de Montr\'eal
    \AND
    \name Golnoosh Farnadi \email farnadig@mila.quebec \\
    \addr Mila, McGill University
}

\def\month{11}  
\def\year{2026} 
\def\openreview{\url{https://openreview.net/forum?id=XXXX}} 

\begin{document}

\maketitle





\begin{abstract}

Despite their widespread success across machine learning, the learning dynamics and generalization properties of neural networks remain poorly understood. 
The neural tangent kernel (NTK) provides an analytical characterization of the learning dynamics of infinitely wide neural networks and serves as a useful proxy for finite-width networks in supervised learning. 
In this work, we extend the NTK framework to reinforcement learning (RL) by introducing a temporal-difference (TD) flow formulation that accommodates time-varying mini-batches and TD targets, yielding a closed-form solution for value function evolution in an overparametrized setting. We also devise an optimization-based state-space exploration method that leverages the closed-form solution, coupled with constraints on the environment, to model the set of feasible states.  On a deterministic grid environment, we validate that the proposed RL-based NTK formulation is indeed learning efficiently and adapting to the distribution shift in the training data distribution, and we demonstrate that the resulting exploration strategy improves sample efficiency.  To our knowledge, this is the first work to extend NTK theory to characterize the learning dynamics of RL training with time-varying mini-batches and TD updates.

\end{abstract}
\section{Introduction}
In the context of reinforcement learning (RL), deep Q-learning marks an important milestone by demonstrating the ability of deep neural networks to learn complex policies~\citep{mnih2015human}. Prior to its introduction, combining tabular methods like Q-learning with nonlinear function approximators was notoriously unstable and prone to divergence. Deep Q-Networks (DQN) resolve these instabilities by introducing experience replay to break temporal correlations and a periodically updated target network to stabilize parameter updates. Its major breakthrough on Atari 2600 games demonstrated that RL could scale directly to high-dimensional visual domains from raw pixels~\citep{mnih2013playing}. Nevertheless, DQN remains susceptible to certain pitfalls in various contexts, such as overestimation bias and sample inefficiency. To address the former, double DQN (DDQN) decouples action selection from value evaluation, using  the online DQN to select the greedy action and the target DQN to evaluate its value during the online network's parameter update. 

In order to better understand how these algorithmic ideas stabilize optimization during training, one can study their training dynamics directly. Observing black-box network updates empirically is often uninformative, making analytical approaches like the neural tangent kernel (NTK) essential. The NTK is a valuable tool to study deep neural networks in simple settings, yet it normally rests on assumptions that are inconsistent with the common RL training framework. Specifically, it is very natural to update parameters using batches of data instead of computing the gradient over the full dataset. In addition, many learning algorithms feature distribution shifts when new data is introduced (e.g., off-policy RL algorithms). 

In this work, our first theoretical contribution is to bridge the gap between classical NTK formulations and RL training introducing a temporal-difference (TD) flow formulation that accommodates time-varying mini-batches and TD targets. For overparameterized ReLU networks (i.e., in the infinite-width limit), we derive a closed-form solution governing value function evolution. On finite state-action spaces, this solution is both strongly convex and smooth. Second, on the practical side, we leverage these properties to formulate a mathematical program, enabling off-the-shelf optimization solvers to search the state-action space globally for high-value exploration targets in a single optimization step. Third, through computational experiments, we validate the proposed NTK formulation by showing that it produces convergent, stable learning behavior, and we demonstrate that our solver-augmented exploration framework substantially improves sample efficiency over standard deep Q-learning baselines.

\section{Literature Review}

\paragraph{Experience replay and mini-batch dynamics}

Replay buffers are used in deep Q-learning to help decorrelate temporal data. Existing methods for memory management rely on empirical tuning, heuristic transformations and statistical approximations over historical samples. We review a few key contributions on the use of replay buffers.

\citet{fedus2020revisiting} provide a detailed ablation study of experience replay (ER) in order to identify key properties that make it efficient. They show that increasing capacity of the replay buffer greatly improves performance in some tasks while not affecting others. Finally, they also characterize the importance of the replay ratio across various tasks. They formulate hypotheses and validate them to obtain general practical recommendations: 1) increasing capacity and decreasing the age of the oldest policy, and 2) using $n$-step returns in an off-policy setting. \citet{liu2018} analyze the properties of the replay buffer on solutions to simple ODEs derived from linear neural networks. They show that the amount of memory can substantially alter the performance of the algorithm and they also derive a regime where prioritized experience replay (PER) is harmful. Our proposed extension of the NTK to multiple mini-batches allows for a continuous-time equivalent of multi-batch training. It can be used to test the same hypotheses in overparameterized networks at a reduced computational cost. Importantly, we devise a solver-augmented exploration strategy that can select state-action pairs with high TD loss under the current model predictions. Contrary to PER,  we do not replay tuples with high TD loss more often, and we treat the added data simply as a new addition to a dataset.

\citet{fujita2025experience} extend random reshuffling (RR) of the dataset to experience replay and show that its theoretical advantages over with-replacement strategies translates to the RL case. \citet{di2022analysis} analyze the theoretical properties of a stochastic process that is pushed in a replay buffer and sampled. Their analysis provides theoretical tools to prove the convergence of multiple replay-based algorithms. The random reshuffling idea can be analyzed using our closed-form solution with mini-batches. While the shuffling ensures that all instances in the buffer are used in training, regardless of the age of the policy used to retrieve them, significantly more weight is attributed to instances that appear multiple times. In a deterministic environment, this can be prevented by only storing unique instances in the buffer; we use this idea in our experiments, and we sample with replacement to satisfy typical SGD assumptions.   

\citet{Isele_Cosgun_2018} explore different selection strategies for experiences that are stored in the buffer. They compare methods that favour surprise, reward, training distribution matching, and state space coverage. They show that training distribution matching is consistently the best approach. However, they derive a case where state space coverage is the best approach. \citet{zhao2024efficient} propose a diversity-based experience replay (DBER) algorithm to achieve state-of-the-art results in sparse result environments. Using the closed-form solution derived via the NTK, we instead use solver-augmented exploration of the state space to select states whose predicted TD loss is high. We note, however, that we can enforce diversity by introducing no-good cuts or hamming cuts to the optimization model that is fed to the solver.

\citet{Andrychowicz2017HindsightER} present a novel algorithm, hindsight experience replay (HER), to improve the sample efficiency of the experience-based algorithms for sparse rewards. The goal of a replayed example is modified post-hoc to obtain a stronger signal during learning. \citet{crowder2024hindsight} apply HER to PPO and demonstrate that it dramatically accelerates the learning. The same can be said of common on-policy learning algorithms on a tested predator-prey model. \citet{kim2023symmetric} introduce symmetric replay training (SRT) in the context of combinatorial optimization problems. Their methodology modifies common RL approaches such as PPO, and even GFlowNets, that seek to obtain high-reward samples. Given a high-reward sample, a symmetric trajectory is replayed uniformly at random. This corresponds to maximizing the entropy of the distribution of trajectories ending in a given state $x$. On all tested approaches, they show improvement in the sample efficiency and overall obtained solutions in the travelling salesman problem (TSP). Finding symmetric trajectories can be computationally hard, and we instead rely on a solver-based approach that leverages the closed-form solution of the NTK in order to select state-action pairs that have high TD loss.

\paragraph{Neural Tangent Kernel Analysis in RL}
Prior RL applications use neural network approximators purely as analytical tools to derive asymptotic global bounds. Next, we revise these results and their relation to the NTK theory.

\citet{xu2020finite} prove that deep Q-learning finds the optimal policy with a $O\left( \frac{1}{\sqrt{T}} \right)$ convergence rate if the network approximator is sufficiently overparameterized (explicitly, wide). Their proof does not rely on the i.i.d. assumption, but requires that the data is generated from an MDP. \citet{qiu2021reward} devise a provably sample and computationally efficient reward-free RL algorithm, making use of nonlinear kernel and neural function approximators for single-agent MDPs. They achieve polynomial sample complexity in $\frac{1}{\epsilon^2}$ for a $\epsilon$-suboptimal policy. \citet{chen2020generalized} show that neural networks trained with noisy gradient descent and weight regularization can exhibit kernel training behaviour. They also obtain generalization bounds using mean-field analysis. They show that kernel approaches, when trained with noisy gradient, can efficiently learn functions if the initial weight distribution is not too far from the optimal weight distribution leading to interpolation on the data. 

\citet{che2024} provide a theoretical analysis of overparameterized linear networks in the context of TD-learning. The authors compare the learning dynamics of various approaches to TD-learning. The first uses the same set of parameters on both sides of the TD equation; the second freezes the right-hand side (i.e. the TD target) and updates the parameters on the left-hand side, periodically updating the target parameters to the latest network parameters. Under mild assumptions, TD-learning with periodic target updates is shown to converge to an optimal solution on the training dataset (offline learning). This work covers linear networks and does not extend to deep neural networks. The results that are provided for Q-learning are in the continuous control case, and while convergence in the parameter space is established, no bounds on the error are given. As a point of comparison in section~\ref{sec:experiments}, we demonstrate the stable learning dynamics of overparameterized deep neural networks under varying mini-batches of data.

\citet{lyle22a} apply the theory of kernel regression in the context of TD learning. Central to their results is a closed-form solution to the gradient flow of a kernel predictor under TD error. The authors argue that the value function evaluated on novel observations depends strongly on the training data, and that the minimization of large TD errors can have a significant interference effect. This observation runs contrary to the intuition that methods that generalize well exhibit weak interference. The gradient flow in the closed-form solution updates both the left-hand and right-hand sides of the TD equation simultaneously; in other words, the TD target does not rely on a frozen instance of the predictor.

\citet{st2025understanding} analyze the properties of the NTK in the lazy regime. They show that, contrary to prior beliefs, the NTK leads to non-trivial predictions. By studying the NTK of infinitely-wide deep ReLU networks, it can be inferred that its eigenvalues are bounded and strictly greater than zero. In the context of stochastic gradient descent analysis, this implies that infinitely-wide deep ReLU networks exhibit both smoothness and strong convexity during training. These two properties combine to yield desirable convergence properties in mini-batch training, when the full dataset is held constant. While it is in theory intractable to train truly overparameterized networks, the existence of a closed-form solution for infinitely-wide deep ReLU networks provides us with a proxy to these models. Moreover, we show in the experiments section how to accommodate the distribution shift in the RL setting when training an NTK model. This approach can also be applied to finite-width neural networks.

\paragraph{Active Exploration and State Space Optimization}


\citet{eysenbach2019search} look at the intersection of RL and planning. They devise a planning algorithm based on the current value estimates. With the obtained goal in mind, they are able to use the RL agent's policy to guide towards that goal. The planning algorithm is run on a graph-based abstraction of the state space obtained via exploration with a RL agent; nodes are derived from observations stored in the replay buffer. However, the current value estimates are given by a black-box and no additional information can be gained. By leveraging the NTK for infinitely-wide deep ReLU networks in our work, we gain access to a closed-form solution, facilitating novel approaches in planning and exploration.

In addition to training a Q-network to maximize cumulative rewards, \citet{simmons2020} train a second neural network whose value function is induce by an auxiliary reward function that is equal to the absolute temporal difference error of the first model. Both models select training data from a shared replay buffer and together, they form a an adversarial training framework where exploration is driven by state-action pairs that lead to high cumulative temporal difference error. Their approach performs comparably or outperforms most baselines on a variety of continuous control tasks. We apply the NTK to this adversarial setup, enabling us to study the learning dynamics in the context of overparameterized networks. The existence of a closed-form solution further allows us to enhance the adversarial exploration by way of constrained optimization problem solvers.

In summary, existing literature addresses mini-batch replay through empirical sample selection, uses NTK theory strictly for asymptotic proofs, and restricts active exploration to local graph rollouts. We bridge these domains by extending NTK gradient flow to sequential mini-batches and temporal difference targets. By leveraging the smoothness and strong convexity of the closed-form NTK proxy \citep{st2025understanding}, we guarantee training stability by choose time increments appropriately, and we replace local exploration heuristics with a mathematical program that optimizes across the feasible state space $\mathcal{C}$ in a single step.

\section{Notation}
In this section, we present the notation and definitions that will be useful for the rest of this article. 

Consider an $L$-layer fully connected neural network parameterized by $\theta=\{W^{(l)}\}_{l=0}^{L-1}$, where $W^{(l)} \in \mathbb{R}^{n_{l+1} \times n_l}$. The input dimension is $n_0$, and $S^{n_0 - 1}$ denotes the $n_0$-dimensional sphere. Let $\alpha^{(0)}=x$ be the input. For a scalar activation function $\sigma: \mathbb{R} \to \mathbb{R}$ applied component-wise, the forward pass is defined recursively for layers $l \in \{1,\ldots,L\}$ as:
\begin{align*}
\tilde{\alpha}^{(l)} &= \frac{1}{\sqrt{n_{l-1}}} W^{(l-1)} \alpha^{(l-1)},
&\alpha^{(l)} &= \sigma\left(\tilde{\alpha}^{(l)}\right)
\end{align*}
where $\tilde{\alpha}(l)$ are the pre-activations, $\alpha(l)$ are the post-activations, and $n_l$ is the width of layer $l$.

\begin{deff}\label{def:hfun}
The function $h: [-1, 1] \to \mathbb{R}$ is defined as
\begin{equation*}
    h(x) = \frac{x \arcsin(x) + \sqrt{1 - x^2}}{\pi} + \frac{x}{2}.
\end{equation*} 
\end{deff}

\begin{deff}\label{def:theta_kernel}
Adopting the notation of~\citet{jacot2018neural}, the functions $\Theta_{\infty}^{(l)}: [-1,1] \to \mathbb{R}$ and $\rho^{(L)}: [-1,1] \to \mathbb{R}$ are defined recursively  with base cases $\rho^{(1)}(x) = x$ and $\Theta_{\infty}^{(1)}(x) = x$  as 
\begin{align*}
    \rho^{(l)}(x) &= h(\rho^{(l-1)}(x)) \\
    \Theta_{\infty}^{(l)} (x) &= \frac{l-1}{l} h'(\rho^{(l-1)}(x)) \Theta_{\infty}^{(l-1)}(x) + \frac{1}{l} \rho^{(l)}(x).
\end{align*}
\end{deff}

Definition~\ref{def:theta_kernel} is a normalized version of the NTK limit kernel from \citet{jacot2018neural}. By abuse of notation, for any $g: [-1,1] \to \mathbb{R}$ and $m \times n$ matrix $A$, $g(A)$ refers to the component-wise application of the function $g$ to the matrix $A$. Moreover, given a dataset $\{x_i \in S^{n_0 -1} \}_{i=1}^n$, the data matrix $X \in \mathbb{R}^{n \times n_0}$ is formed by stacking $x_i^\top$ along its rows. This yields the kernel matrices $\rho^{(l)}( X X^\top)$ and $\Theta_{\infty}^{(l)} (X X^\top )$. For parameters $\theta = \{W^{(l)} \}_{l=0}^{L-1}$, the notation $\theta_t$ refers to the parameters at time $t \geq 0$. When clear from the context, any time-dependent mathematical object with subscript $t$ refers to that object evaluated at time $t$.

\section{Mini-batch training with the NTK}\label{sec:minibatch}

The typical setting for training with the NTK, namely NTK regression, uses the same dataset over the full training horizon. By exploiting the closed-form solution to the gradient flow dynamics~\citep{jacot2018neural, arora2019exact}, one can obtain the output formula for arbitrary time $t$.  Crucially, this solution implicitly assumes that the gradient is computed over the entire dataset simultaneously. However, practical neural network training updates parameters sequentially using batches of the dataset. Thus, when adapting this sequential updating procedure to its gradient flow equivalent, we can derive the output of an infinitely-wide neural network trained on successive data batches. Mathematically, given indices $b \in \{1, \dots, B\}$ of dataset batches, where $B \in \mathbb{N}$, we formalize the evolution of the model output $f_t$ via gradient flow and changing data batches in Definition~\ref{def:diffeq}.\footnote{This is not a strictly continuous gradient flow over the entire horizon, as the time derivative exhibits jump discontinuities at the batch transition times $t_b$.}

\begin{deff}\label{def:diffeq}
    Given a time horizon $T \in \mathbb{R}_{> 0}$ and a batch of data $\{X^{(b)}\}_{b \in [B] }$ where $[B] = \{1, \dots, B \}$, we call a sequence of stopping times any sequence of real numbers $0 = t_1 < \dots < t_B = T$ such that the NTK at time $t \in [t_b, t_{b+1} [, b \in [B-1]$ evolves with respect to $X^{(b)}$. Specifically, the dynamics of the gradient flow are given by
    \begin{align*}
        \partial_t f_t(x) &= -\Theta_{\infty}^{(L)} \left( x^\top X^{(b)\top} \right) \left( f_{t} \left( X^{(b)} \right) - y^*\left( X^{(b)} \right) \right), \\
        t &\in [t_b, t_{b+1}[, b \in [B-1]
    \end{align*}
    where $y^*\left(X^{(b)} \right)$ are the target values over the batch $X^{(b)}$.
\end{deff}

In Section~\ref{sec:application_RL}, we provide Proposition~\ref{prop:theta_bounded_constant}, which shows that in the infinite-width limit, the NTK (i.e. $\Theta_{\infty}^{(L)}$) does not change when replacing data batches at fixed intervals. As in Definition~\ref{def:diffeq}, the dynamics evolve with respect to the latest batch $X^{(b)}$, however the kernel function $\Theta^{(L)}_t$ does not vary (i.e., is independent of $t$). Therefore, even with a constant kernel, the specific ordering and selection of data batches significantly influences the final output of the model at the end of the horizon (time $T$). We underline that while standard supervised mini-batch flow assumes fixed targets $y^*$, RL introduces non-stationary TD targets by default. However, in alignment with common learning approaches, we fix the TD targets during training. We describe the resulting dynamics, which we refer to as the Temporal Difference Flow formulation, in Section~\ref{sec:application_RL} to extend the NTK framework to RL.

\section{Extension to RL}\label{sec:application_RL}

A stated goal of our analysis of the NTK is to offer a more realistic proxy to neural network training in order to prototype, benchmark, or even understand various RL algorithms. Prior work focuses on a fixed dataset throughout training and the differential equation underlying the learning dynamics is taken to the infinite time limit. In practice, these assumptions are not met and to this end, we endeavour to adapt the NTK analysis to a setting with varying batches of data and a finite time training schedule. Meanwhile, a realistic proxy in the RL setting must incorporate the method by which TD targets are updated over time. In light of these requirements, we provide a differential equation and its corresponding solution that satisfies the sought-after framework.

This section details key aspects of the NTK in the context of RL. First, in the infinite-width regime, the kernel stays constant throughout training, as highlighted in Proposition~\ref{prop:theta_bounded_constant}. Second, the training dynamics follow a specific differential equation given by Definition~\ref{deff:tdf}, from which we obtain a closed-form solution in Theorem~\ref{thm:td_closed_formulation}. Third, the kernel is shown to exhibit invertibility and induce certain smoothness properties in the training objective. Finally, we provide an iterative procedure to compute the closed-form solution of Theorem~\ref{thm:td_closed_formulation} for arbitrary inputs, and we analyze the computational cost derived from training with varying mini-batches and performing early stopping.

\paragraph{Preliminaries}
In the RL setting, we have to adapt the aforementioned sequential NTK framework to account for the different nature of the temporal difference objective that we optimize. In this work, we suppose that we have a deterministic environment. In other words, given a state $s$ and an action $a$, we can compute the next state $s'$ with probability $1$. In contrast to the typical learning framework introduced in the previous section, the data points come in the form of tuples $(s, a, s', r)$, where $s$ is the state prior to the action selection, $a$ is the action taken, $s'$ is the state obtained by transitioning from $s$ after taking action $a$, and $r$ is the corresponding reward depending on $(s,a)$. We can relate this to the typical learning notation of data $(x, y)$ by setting $x = (s, a)$ and $y = r$. Here, the vector $x$ corresponds to the features of the data point, and $y$ acts as the label. The value $\gamma \in \left]0, 1\right]$ is the discount factor, while the parameterized action-value network is $Q_{\theta_t}$. Given a target network frozen at time $\hat{t}$, the objective is thus to minimize the Euclidean 2-norm error for $t \geq \hat{t}$: $\left\lVert Q_{\theta_t}(s, a) - r(s, a) - \gamma \max_{a'} Q_{\theta_{\hat{t}}} (s', a') \right\rVert_2$. 

\paragraph{Constant NTK for RL} 
Just as in the supervised setting, taking the infinite-width limit of these overparameterized networks allows us to recover a kernel formulation. Specifically, the NTK converges to a limiting kernel as the hidden layer widths go to infinity as shown next in Proposition~\ref{prop:theta_bounded_constant}.

\begin{prop}\label{prop:theta_bounded_constant}
    Let $d_t: \mathbb{R}^{n_0} \to \mathbb{R}^{n_L}$ be such that $\int_{0}^T \lVert d_t \rVert_X < \infty$ with probability $1$ for some $T > 0$ and let $\sigma$ be $c$-Lipschitz. The kernel $\Theta_{t}^{(L)}$ defined by
    \begin{equation*}
        \Theta^{(L)}_t(x, x')= \nabla_{\theta_t}Q_{\theta_t}(x )^\top \nabla_{\theta_t}Q_{\theta_t}(x' )
    \end{equation*}
    is bounded on $\left[ 0, T \right]$ in the limit $n_1, \dots, n_{L-1} \to \infty$. Moreover, it remains constant on the interval $[0, T]$. 
\end{prop}

The proof of Proposition~\ref{prop:theta_bounded_constant} follows a similar treatment as Theorem 2 in \citet{jacot2018neural} and is provided in Appendix~\ref{sec:appendix_prop_proof}. Building upon Proposition~\ref{prop:theta_bounded_constant}, we show that given a sequence of mini-batches, it is possible to obtain a closed-form solution by following the gradient flow equation on each time interval. 

\paragraph{Flow of the Q-value function}
Before formalizing the main statement of this section as Theorem~\ref{thm:td_closed_formulation}, we introduce some useful definitions.

\begin{deff}[Inner product space induced by $X$]\label{deff:inner_product_space}
    Let $X_{i,\cdot} = x_i^\top$ for $i=1,\dots,n$. The inner product space induced by $X$ is the space of functions $\mathbb{R}^{n_0} \to \mathbb{R}^{n_L}$ such that
    \begin{equation*}
        \left\langle f, g \right\rangle = \frac{1}{n} \sum_{i=1}^n f(x_i)^\top g(x_i).
    \end{equation*}
    The notation $\left\lVert f \right\rVert_X$ refers to $\sqrt{\langle f , f \rangle}$.
\end{deff}

\begin{deff}
    The function $\Phi_{\kappa}$ given a kernel $\kappa: \mathbb{R}^{n_0 \times n_0} \to \mathbb{R}^{n_L}$ on a dataset $X$ of size $n$ is defined by
    \begin{equation*}
        \Phi_{\kappa}(\delta) := \delta(\kappa\left(x, \cdot\right)),
    \end{equation*}
    where $\delta$ is any dual function with respect to the inner product space induced by the dataset $X$. 
\end{deff}

\begin{deff}
    Let $X_{i, \cdot} = x_i^\top$ for $i=1,\dots,n$. Let $\mathcal{N}(x)$ be the neighbourhood of $x$, i.e., the admissible states that can be reached in one step from $x$, and define the function $\mu_t(x) \in \mathcal{N}(x)$ via
    \begin{equation*}
        \mu_t(x) = \argmax_{x' \in \mathcal{N}(x)} Q_t(x')
    \end{equation*}
    and define 
    \begin{equation*}
        \mu_t\left(X\right)_{i} = \mu_t(x_i)
    \end{equation*}
    over the dataset $X$ by abuse of notation.
\end{deff}

With the function space and neighborhood mechanics in place, we can now define the exact dynamics of the network under our batch-trained NTK framework.

\begin{deff}[Temporal difference flow equation (TDF)]\label{deff:tdf}
    Given a sequence of stopping times $\{t_1, \dots, t_B\}$ and $\hat{t}_b \in \{t_1, \dots, t_b \}$ for all $b \in [B-1]$,
    the solution $Q_t$ to the flow equation for $t \in [t_b, t_{b+1}[$ is given by
    \begin{align*}
        \partial_t Q_t(x)  &= - \left( \Phi_{\Theta_t^{(L)}}(\delta_t) \right),
    \end{align*}
    where $\delta_t = \langle d_t, \cdot \rangle$ and $d_t = Q_t(X^{(b)}) - \gamma Q_{\hat{t}_b} \circ \mu_{\hat{t}_b}(X^{(b)}) - R(X^{(b)})$, is called a solution to the temporal difference flow equation (TDF). 
\end{deff}

By solving the equation in Defintion~\ref{deff:tdf}, we obtain a closed-form solution that gives a proxy to the Q-value approximator of an overparameterized neural network in Theorem~\ref{thm:td_closed_formulation}. In order to simplify notation in the Theorem, we first introduce the following definition:
\begin{deff}\label{def:exponential_matrix}
    We define the function $\Lambda_{s,t}^{(L)}: \mathbb{R}^{n \times n_0} \to \mathbb{R}^{n \times n} $ as
    \begin{equation*}
        \Lambda_{s,t}^{(L)}(X) =  \left( \exp\left( -(t-s) \Theta_s^{(L)} \left(X X^\top \right) \right) - I_n\right).
    \end{equation*}
\end{deff}

Definition~\ref{def:exponential_matrix} is used to compute the closed-form solution of $Q_t$ on a sub-interval $[t_b, t_{b+1}[$ via Theorem~\ref{thm:td_closed_formulation}.

\begin{thm}\label{thm:td_closed_formulation}
    Given a sequence of stopping times $\{t_1, \dots, t_B\}$ and $Q_0$, the function $Q_t$ has a closed-form solution $\mathbb{R}^{n_0} \to \mathbb{R}$ given by the solution to the TDF (Definition~\ref{deff:tdf}) on each interval $[t_i, t_{i+1}[$ for $i \in [B-1]$. Given $\hat{t}_b \in \{t_1, \dots,t _b\}$ and an invertible $\Theta_{\infty}^{(L)} \left( X X^\top \right)$, the closed-form solution on $t \in [t_b, t_{b+1}[$ is given by
    \begin{align*}
    Q_t(x) &= \Theta^{(L)}_t (x^\top X^{(b)\top}) \left( \Theta_t^{(L)} \left( X^{(b)} X^{(b)\top} \right) \right)^{-1} \\
    &\Lambda_{t_b,t}^{(L)}(X^{(b)})
    \left( Q_{t_b}(X^{(b)}) - R(X^{(b)}) - \gamma Q_{\hat{t}_b} \circ \mu_{\hat{t}_b} (X^{(b)}) \right) \\
    &+ Q_{t_b} ( x ),
\end{align*}
    where $\Theta_t^{(L)} = \Theta_{\infty}^{(L)}$ for all $t \geq 0$. 
\end{thm}

The time $\hat{t}$ refers to a prior batch where the target network was frozen (e.g., see \citet{mnih2015human}). The closed-form solution of Theorem~\ref{thm:td_closed_formulation} evolves with kernel $\Theta_{\infty}^{(L)}\left( X^{(b)} X^{(b)\top} \right)$ on the interval $[t_b, t_{b+1} [$. By comparison with the supervised learning setting, Theorem~\ref{thm:td_closed_formulation} follows similar training dynamics within each intervals $[t_b, t_{b+1}[$; in contrast, the evolution of the solution sews together various pieces in a non-smooth continuous path. In that sense, the solution to to TDF is not a true gradient flow solution, but rather a right-derivative counterpart since the right-hand derivative of the solution exists at each discontinuity point $t$, while it is in fact differentiable everywhere else. Below, we detail properties of the NTK that are useful in theoretical analysis. 

\paragraph{Properties of $\Theta_{\infty}^{(L)}$}
We can show that the continuous-time TD error is $n^{(b)}$-smooth with respect to $t$ along any interval $[t_b, t_{b+1}[$, with $n^{(b)} = |X^{(b)}|$ since the diagonal of $\Theta_{\infty}^{(L)} \left( X^{(b)} X^{(b)\top} \right)$ consists of ones. It is also $\chi$-strongly convex with respect to $t$ on that interval if the smallest eigenvalue of $\Theta_{\infty}^{(L)} \left( X^{(b)} X^{(b)\top} \right)$ is $\chi > 0$. 
The following proposition (Proposition~\ref{prop:theta_positive_definite}) states that for a depth $L \geq 2$, the NTK is a positive definite kernel. In particular, ReLU activation and fully-connected networks yield a closed-form solution that is strongly convex and smooth on each interval.

\begin{prop}[from \citet{jacot2018neural}]\label{prop:theta_positive_definite}
    The NTK $\Theta^{(L)}_t$ is positive definite with probability $1$ for $L \geq 2$. 
\end{prop}

\paragraph{Computation of closed-form solution}
When computing the closed-form solution in Theorem~\ref{thm:td_closed_formulation}, by inspection of the equation we can compute $Q_t(x)$ by recursively compute $Q_{t_b}$ are prior stopping times. This yields the following decomposition:

\begin{align*}
    u(x)_b^\top &:= \Theta_{\infty}^{(L)} \left( x^\top X^{(b)\top} \right) \\
    v_b &:= \left( \Theta_{\infty}^{(L)} \left( X^{(b)} X^{(b)\top} \right) \right)^{-1} \Lambda_{t_b,t}^{(L)} \left( X^{(b)} \right) \\
    &\left( Q_{t_b} \left( X^{(b)} \right) - R \left( X^{(b)} \right) - \gamma Q_{\hat{t}_b} \circ \mu_{\hat{t}_b}\left( X^{(b)} \right) \right),
\end{align*}

from which $Q_t(x) = \sum_{1, \dots, b \mid t_b \leq t} u(x)_b^\top v_b$. The values $v_b$ are kept in memory as they are computed, and for a new batch $X^{(b)}$ of data, the values $u(x_i)_b$ are computed for $x_i \in X^{(b)}$. Each new $u(x_i)_b$ is used to compute the new term $v_b$ that is stored in memory. The time spent to compute $Q_t(x)$ is $O \left( g^2(t) \right)$, where $g(t) = | \{ t_i \mid t_i \leq t \} |$\footnote{The batch size $n_b$ and depth $L$ are held constant.}. This is much more efficient than training a deep neural network with extremely wide hidden layers, given the condition that their width is much larger than the number of data points in the replay buffer.

\section{Solver-augmented exploration}\label{sec:solver_exploration}
The analytical proxy for neural network training dynamics that is described in section~\ref{sec:application_RL} provides a concise mathematical tool that can be queried for insights into its predictions. Deterministic environments that are suitable to description via constraints pave the way to new avenues for exploration in RL; we exploit the solution from Theorem~\ref{thm:td_closed_formulation} and formulate a search problem to gather new states that are stored in a buffer for subsequent learning.


Our algorithmic approach searches for a solution $x \in \mathcal{C}$, where $\mathcal{C}$ is the set induced by constraints on the state-action space $S \times A$ and the variable $x$ corresponds to the description of a state-action pair $x = (s, a)$. In a grid world example, the variable $x$ can be decomposed into variables $\hat{s}$ and $\hat{a}$: $\hat{s} \in \mathbb{R}^{h \times w}$ (for a grid of size $h \times w$) models each possible cell content, with different cases equivalent to different integers, while $\hat{a} \in \mathbb{R}$ indexes the different action scenarios. Constraints are introduced to ensure that $\hat{s}$ describes a a valid state. Given $t_b$, the objective is to find $x^* \in \mathcal{C}$ such that the estimate $Q_{\infty}(x^*)$ is maximized or minimized. This objective contributes to surprise whenever an optimal $x^* \in \mathcal{C}$ is not part of the buffer and the TD target of $x^*$ is not close to the estimate. In this case, the error is corrected during the next training step. We formulate the search problem \ref{eq:nearconvex_program}, which is passed to a MINLP solver to find $\argmax_{x \in \mathcal{C}} Q_{\infty}(x)$\footnote{In the case of maximization.}:

\begin{align*}\label{eq:nearconvex_program}
    \min_{x,y,u} &\mp c_{\omega}^\top u  \tag{$P^{\pm}_{\omega}(\lambda)$} \quad \text{subject to} \\
    y &= X x \\
    \Theta_{\infty}^{(L)} \left( X X^\top \right) u &=  \Theta_{\infty}^{(L)} (y)\\
    c_{\omega}^\top u &\in \mathbb{R}^{\pm} \\
    x &\in \mathcal{C},
\end{align*}
where the cost $c_w$ corresponds to the TD target under the sample $\omega$ with initial value $R\left( X \right) + \gamma Q_0 \circ \mu_0 \left(X \right)(\omega)$. With \eqref{eq:nearconvex_program}, it is possible to recover a solution $x \in \mathcal{C}$ such that the Q-value is either maximized or minimized ($\pm$). The random variable $R\left( X \right) + \gamma Q_0 \circ \mu_0 \left( X \right)$ can also be replaced by its mean $R\left(X^{(1)} \right)$ as is the case in section~\ref{sec:experiments}. 

The closed-form estimate $Q_{\infty}$ is a proxy for a single batch update as we take $t \to \infty$ while holding the mini-batch constant for the interval $[0, \infty[$, achieving perfect interpolation on the new mini-batch. Alternatively, we could model $Q_t$ directly via the equation of Theorem~\ref{thm:td_closed_formulation} by introducing additional constraints for $t_1, \dots, t_B$ and the corresponding mini-batches $X^{(1)}, \dots, X^{(B)}$. For simplicity and to make \ref{eq:nearconvex_program} more compact, we focus on the first use case. We observe that $\Theta^{(L)}_{\infty}$ is a convex function component-wise; we handle the nonconvex constraints using a mixed-integer nonlinear programming (MINLP) solver running a specialized branch-and-bound algorithm~\citep{vigerske2018scip}. With the mathematical program \ref{eq:nearconvex_program} now defined, we introduce Algorithm~\ref{alg:nearconvex_program}.

\begin{algorithm}
    \begin{algorithmic}
        \State \textbf{Input}: Dataset $X$, small $\lambda > 0$, $m \in \mathbb{N}$ 
        \State Draw $\omega$ at random
        \State Draw $+$ or $-$ case at random
        \For{$m$ iterations}
        \State Solve $P^{\pm}_{\omega}(\lambda)$ with $X$
        \State Recover $x^* \in \mathcal{C}$ and observe $R(x^*)$
        \If{TD error is high}
        \State $X \gets X \cup \{x^*\}$
        \EndIf
        \EndFor
        \State \Return $X, \omega$, $\pm$ case
    \end{algorithmic}
    \caption{Solver-Augmented State-Space Exploration}
    \label{alg:nearconvex_program}
\end{algorithm}

Using the output of Algorithm~\ref{alg:nearconvex_program}, we can augment the initial dataset $X$ to define a new generalizing function. Since at each iteration in the loop we augment the dataset with $x \in \mathcal{C}$ by selecting the point with the highest Q-value, we obtain a sequence of high predicted value candidates across the full set $\mathcal{C}$. When solving $P^{\pm}_{\omega}$ and once an optimal $x^*$ is found, the dataset $X$ can be extended with $x^*$. If Algorithm~\ref{alg:nearconvex_program} is run with $m = \left\lvert \mathcal{C} \setminus X \right\rvert$, the final model would indeed interpolate perfectly the whole state space as $t \to \infty$. While conceivable in theory, this is practically not achievable. However, the model gains a certain amount of surprise from the procedure at each step, as it finds the $x$ believed to have highest value. 

Since the condition number of $\Theta_{\infty}^{(L)} \left( X X^\top \right)$ can be high for large $n$, it can be useful to make a change of variable with $\delta_\omega$ and rewrite the objective $c_{\omega}^\top u$ as $d_{\omega}^\top \Theta_L$ after solving
\begin{equation}\label{eq:constr_change_of_var}
    \Theta_{\infty}^{(L)} \left( X X^\top \right) d_{\omega} = c_{\omega}
\end{equation}
and removing the constraint $\Theta_{\infty}^{(L)} \left( X X^\top \right) u = \Theta_\infty^{(L)} (y)$. Expression~\eqref{eq:constr_change_of_var} only involves the variable $d_\omega$ and if the equation is solved beforehand, the changed objective is dependent only on $\Theta_{\infty}^{(L)}(y)$ \footnote{This term is modelled with variable $\Theta_L$ in \ref{sec:kernel_constraints}.}, for which both lower and upper bounds exist. Therefore, spatial branch-and-bound techniques will be able to exploit the restricted domains in their search. This simplification comes at the cost of solving \eqref{eq:constr_change_of_var} for $d_\omega$, which is $O(n^3)$ via Cholesky decomposition for a dense matrix. There also exist alternatives like, GMRES-based iterative refinement~\citep{carson2018accelerating} or using higher precision floating-point numbers, that return a solution to \eqref{eq:constr_change_of_var} with high precision when the condition number is large.

\section{Experiments}\label{sec:experiments}
In this section, we perform experiments to demonstrate the usefulness of our NTK-based approach to model common RL algorithms. We focus on deep Q-learning, the simplest buffer-based algorithm whose update is modelled directly by the TD flow of Definition~\ref{deff:tdf}, but our method extends to other buffer-based algorithms. Prior work has already compared NTK-based approaches with neural networks~\citep{st2025understanding, Lee2020, yang2020function} and highlighted the similitudes and differences between the two. In order to bridge the divide, we endeavour to adjust the typical NTK framework to align more closely with empirical practice and study its behaviour directly. We divide the set of experiments along two axes. In the first axis, we study the effect of training on a sequence of mini-batches, rather than on a fixed dataset, where the mini-batches are drawn from a replay buffer whose contents drift as new data is collected. In the second axis, we study the sample efficiency of introducing a MINLP solver (Gurobi 13) that optimizes the closed-form NTK estimate of the Q-value in the exploration of the state-action space (i.e., Algorithm~\ref{alg:nearconvex_program}). All our experiments involve the deterministic environment \texttt{DoorKey} from the Minigrid library~\citep{MinigridMiniworld23}. This choice of environment satisfies the assumptions of our closed-form solution, namely deterministic transitions and a finite state-action space; its combinatorial structure, where the goal is reachable only after picking up a key and unlocking a door, makes exploration nontrivial. In order to alleviate the issue of the reward sparsity, we implemented a customized version of the environment, \texttt{DenseDoorKey}, which introduces rewards for subgoals. Details are provided in Appendix~\ref{sec:dense_doorkey}. 

\subsection{Training with minibatch sampling}
\begin{figure}
    \centering
    \includegraphics[width=0.98\linewidth]{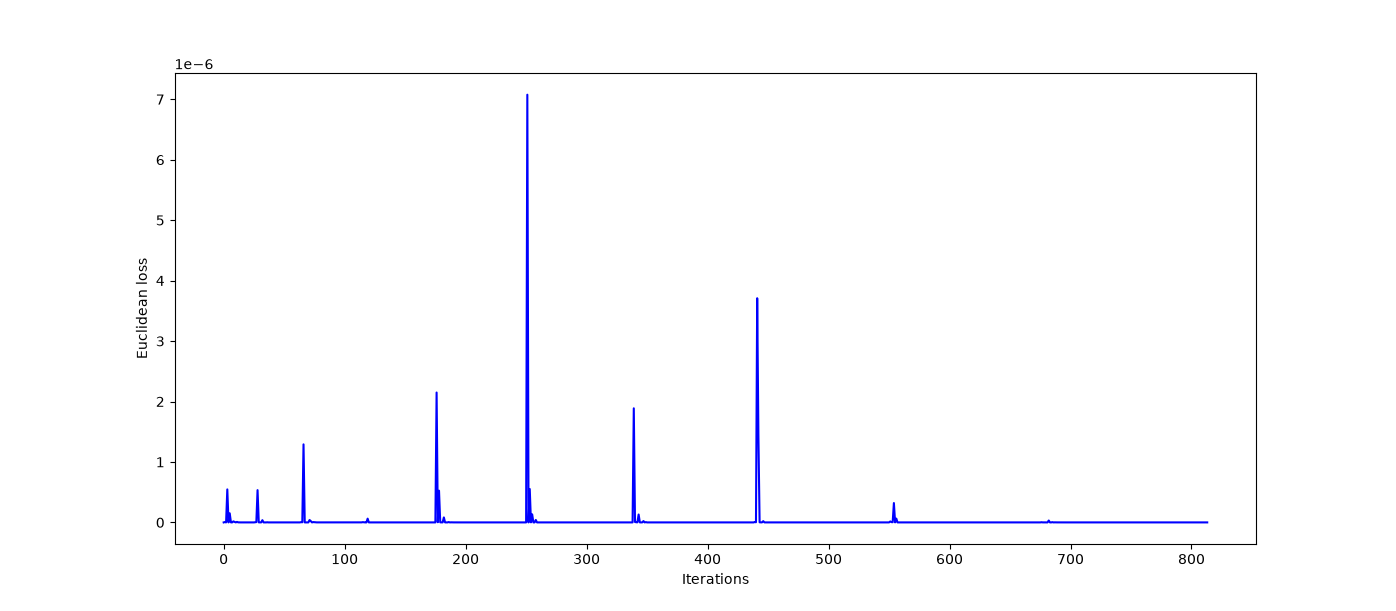}
    \caption{Training loss of the gradient flow equivalent to deep Q-learning (i.e. the NTK). Each iteration corresponds to a training step, i.e. the closed-form solution is updated via $K$ mini-batches selected from the fixed replay buffer.}
    \label{fig:training_loss}
\end{figure}

To evaluate the learning capability of the NTK approach, we study the training loss (i.e. Euclidean loss) as a function of the number of training steps for a deep Q-learning-type algorithm on the \texttt{DenseDoorKey} environment, where the closed-form solution of Theorem~\ref{thm:td_closed_formulation} replaces the network as the action-value estimator. In order to stabilize training, we collect new data and then hold the replay buffer fixed while performing $K$ mini-batch training steps, before resuming data collection; we repeat these two steps throughout training. The time increments for each batch are decreased using a typical stochastic gradient descent schedule with $O\left(\frac{1}{\sqrt{b}}\right)$. When introducing fresh data, the schedule is restarted to prevent training collapse. Moreover, the target Q-values are kept frozen and updated periodically as in regular DQN. We plot the loss in Figure~\ref{fig:training_loss}, where we observe that the training loss peaks after introducing new data, but that it quickly falls back to near-zero. 

The two-step process and the scheduling idea are inspired by regular SGD training, since on a fixed dataset (or batch), the Euclidean loss of the output induced by the NTK of dimension $n \times n$ is both $n$-smooth and $\chi$-strongly convex for some $\chi > 0$. On a finite, yet large state-action space, each batch corresponds to a $\chi_b$ parameter and $\chi$ can be taken to be $\chi = \min_{b \in [B]} \chi_b > 0$. Figure~\ref{fig:training_loss} shows that the Q-values are adapting to the TD targets.

\subsection{Solver-augmented exploration}
In this section, we take advantage of the closed-form solution of infinitely-wide deep ReLU networks to the exploration of the deterministic grid environment. Given that the \texttt{DoorKey} environment can be described by constraints and integer variables, we can use the solver-augmented exploration approach of section~\ref{sec:solver_exploration} to select state-action pairs with high value. We provide additional details in Appendix~\ref{sec:solver_model}. During the exploration of the state-action space, we call Algorithm~\ref{alg:nearconvex_program} as a callback to obtain state-action pairs that are added to the replay buffer. We used Gurobi 13, leveraging its nonlinear mixed-integer solving capabilities. In order to evaluate the sample efficiency of this approach, we compare off-policy RL algorithms with and without solver callback. In addition to DQN, we use a softmax version of the Bellman operator when rolling out the policy. We refer to this approach as SQN. All algorithms use the standard hyperparameters from the \texttt{stable\_baselines3} codebase~\citep{stable-baselines3}. The solver is queried as a callback after each run of the exploratory policy. If a new state-action pair is discovered, it is added to the replay buffer and used in subsequent calls to the solver to build the internal mathematical model. We perform a total of 1000 learning steps and report the rollout policy score in Table~\ref{tab:sample_efficiency}. The rollout score is computed by evaluating total trajectory rewards using the learned policy.

\begin{table}[htb]
    \centering
    \begin{tabular}{c|c}
         \textbf{Algorithm} & \textbf{Rollout score} \\
         \hline
         DQN & 0.0 (0.0)\\
         DQN + solver & 0.10 (0.07)\\
         SQN &  0.74 (0.24)\\
         SQN + solver & 1.02 (0.22)\\
    \end{tabular}
    \caption{Average (std) rollout scores on various RL algorithms on 100 episodes in a $8 \times 8$ grid.}
    \label{tab:sample_efficiency}
\end{table}

\begin{figure}[htb]
    \centering
    \includegraphics[width=0.7\linewidth]{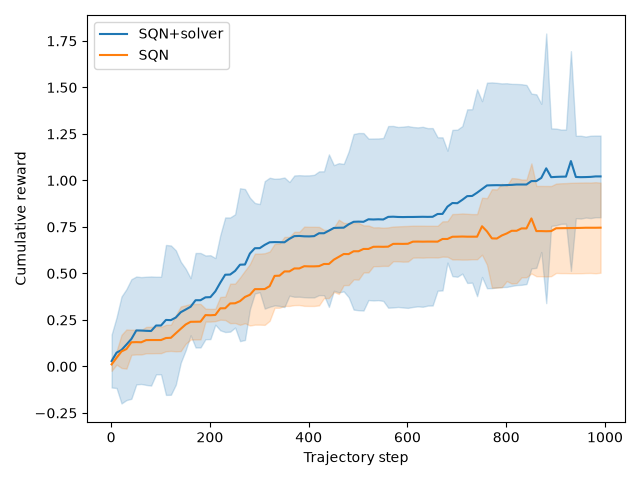}
    \caption{Average rollout scores for SQN and SQN + solver after 1000 learning steps in a $8 \times 8$ grid. The standard deviations are computed from 10 training runs for each method.}
    \label{fig:cum_rewards}
\end{figure}

In both cases, introducing the solver improves the rollout score. To complement Table~\ref{tab:sample_efficiency}, we compare policy evaluation curves of the best performing approach, SQN, with its solver-augmented version during rollouts in Figure~\ref{fig:cum_rewards}. We used different training instances than in Table~\ref{tab:sample_efficiency}. We observe a steady increase in the performance as we perform more learning steps in both approaches, yet the addition of the solver improves the sample efficiency. While these results cannot be expected in all environments, the ability to search for high valued inputs can guide the exploration to new areas of the state-action space that are profitable. In addition, it becomes much easier to favour exploration by adding constraints that forbid the solver from returning state-action pairs it has already encountered (no-good cuts), or that force the next solution to differ from previous ones in a prescribed number of components (Hamming cuts). These constraints let us explore regions of $\mathcal{C}$ far from the solutions previously found. In problem instances where diversity is key, this capability is especially useful. 

\section{Conclusion}

In this paper, we show how to apply the NTK to a reinforcement learning setting, allowing us to bridge the rich theory of overparameterized deep neural networks with off-policy learning algorithms like deep Q-learning. We demonstrate the ability to extend the NTK to varying batches of data that better align with the common practice of neural network training. Through the closed-form solution of infinitely-wide deep ReLU networks, we can also stabilize the training by satisfying typical convergence criteria in stochastic gradient descent, in combination with delayed data collection. Our approach serves as an introduction to the exploring the use of mini-batches and finite time evaluation in overparameterized neural networks. These ideas can be applied to other RL algorithms. As a future research direction, we highlight the possibility of understanding various buffer replay mechanisms and state exploration methodologies by studying the closed-form solution of the NTK. In addition, our solver-augmented exploration approach leaves a rich set of possibilities to be explored; it would be of particular interest to understand how to balance diversity with high value candidates during exploration. Moreover, a central idea is to understand how surprise affects the model predictions going forward.

\bibliography{book/ref}
\bibliographystyle{tmlr}

\newpage

\appendix

\section{Proof of Proposition~\protect{\ref{prop:theta_bounded_constant}}}\label{sec:appendix_prop_proof}
To prove Proposition~\ref{prop:theta_bounded_constant}, we require a generalized form of Grönwall's lemma.

\begin{lem}[Generalized Grönwall's lemma]\label{lem:gronwall}
    Let $u, v \in C^1(]a,b[) \cap C([a,b])$ and $F \in C^1(\mathbb{R})$ with $a<b$. Suppose that
    \begin{align*}
        u'(t) &\leq F(u(t)) \quad t \in ]a, b[ \\
        v'(t) &= F(v(t)) \quad t \in ]a, b[ \\
        v(0) &= u(0).
    \end{align*}
    Then, $u(t)$ satisfies
    \begin{equation*}
        u(t) \leq v(t) \quad t \in ]a, b[.
    \end{equation*}
\end{lem}

We can now prove Proposition~\ref{prop:theta_bounded_constant}. The proof strategy closely follows the proof of Theorem 2 from \citet{jacot2018neural} and we make use of implicit index summation to simplify notation. 

\begin{proof}[Proof of Proposition~\ref{prop:theta_bounded_constant}]
    We prove the statement for $L=2$ and the extension to arbitrary depth $L > 2$ follows by induction over layers. First, we have
    \begin{align*}
        \partial_t \tilde{\alpha}^{(l)} &= \Phi_{\Theta^{(l)}_t}\left( \langle d_t^{(l)}, \cdot \rangle \right) \\
        \partial_t W^{(l)} &= \frac{1}{\sqrt{n_l}} \langle \alpha^{(l)} , d_t^{(l+1)} \rangle,
    \end{align*}
    where 
    \begin{equation*}
        d_t^{(l)} =
        \begin{cases}
            d_t &\text{if } l = L \\
            \left(\sigma'\left(\tilde{\alpha}^{(l)}\right)  \right)
            \odot 
            \left(\left( \frac{1}{\sqrt{n_l}} W^{(l)} \right)^\top d_t^{(l+1)} \right) &\text{if } l < L.
        \end{cases}
    \end{equation*}
    The subnetwork NTKs are
    \begin{align*}
        \Theta^{(1)}_{t} &= \frac{1}{n_0} \left( \alpha^{(0)}  \right)^\top \left( \alpha^{(0)}  \right) \otimes I_{n_1} \\
        \Theta^{(2)}_t &= \frac{1}{n_1} W^{(1)} \sigma'\left( \tilde{\alpha}^{(1)} \right) \Theta^{(1)}_t 
        \sigma'\left( \tilde{\alpha}^{(1)} \right)  W^{(1)}\\
        &+ \frac{1}{n_1} \left( \alpha^{(1)} \right)^\top \left( \alpha^{(1)}  \right).
    \end{align*}
    We set 
    \begin{align*}
        w^{(l)}(t) &= \left\lVert \frac{1}{\sqrt{n_l}} W^{(l)}(t) \right\rVert_2 \\
        a^{(l)}(t) &= \left\lVert \frac{1}{\sqrt{n_l}} \alpha^{(l)}(t) \right \rVert_X 
    \end{align*}
    and we obtain the bounds
    \begin{align*}
        \left\lVert d_t^{(1)} \right\rVert_X &\leq c w^{(1)}(t) \left\lVert d_t^{(2)} \right \rVert_X \\
        &= c  w^{(1)}(t) \left\lVert d_t \right\rVert_X \\
        \left\lVert \Theta^{(1)}_t \right\rVert_2 &\leq (a^{(0)}(t))^2 \\
        \left \lVert \Theta^{(2)}_t \right\rVert_2 &\leq c^2  (w^{(1)}(t))^2 \left\lVert \Theta^{(1)}_t \right \rVert_2 + (a^{(1)}(t))^2.
    \end{align*}
    We set
    \begin{align*}
        \Delta_a^{(l)}(t) &= \left\lVert \frac{1}{\sqrt{n_l}} \left( \tilde{\alpha}^{(l)}(t)  - \tilde{\alpha}^{(l)}(0)  \right) \right \rVert_X \\
        \Delta_w^{(l)}(t) &= \left \lVert \frac{1}{\sqrt{n_l}} \left( W^{(l)}(t) - W^{(l)}(0) \right) \right \rVert_2 \\
        A(t) &= a^{(1)}(0) + \Delta^{(1)}_a(t) + w^{(0)}(0) + \Delta^{(0)}_w(t) \\
        &+ a^{(2)}(0) + \Delta^{(2)}_a(t) + w^{(1)}(0) + \Delta_w^{(1)}(t).
    \end{align*}
    From these last definitions, we obtain the bounds
    \begin{align*}
        \partial_t \Delta_a^{(l)}(t) &\leq \frac{1}{\sqrt{n_l}} \left \lVert \Theta^{(l)}_t \right \rVert_2 \left \lVert d^{(l)}_t \right \rVert_X \\
        \partial_t \Delta_w^{(l)}(t) &\leq \frac{1}{\sqrt{n_l}} a^{(l)}(t) \left\lVert d_t^{(l+1)} \right \rVert_X.
    \end{align*}
    We use these bounds to bound the derivative $\partial_t A(t)$ with
    \begin{equation*}
        \partial_t A(t) \leq \frac{1}{\sqrt{n_1}} q\left(w^{(0)}(t), w^{(1)}(t), a^{(1)}(t), a^{(2)}(t)\right) \left \lVert d_t \right \rVert_X
    \end{equation*}
    for some polynomial $q$ with positive coefficients. Observe that $A(0)$ is stochastically bounded as $n_1 \to \infty$. Using Lemma~\ref{lem:gronwall} and using the fact that $q$ is bounded and continuous over $\left[0, T\right]$, we obtain $A(t)$ uniformly bounded over $\left[ 0, T \right]$. This implies that $\partial_t A(t) \to 0 $ as $n_1 \to \infty$, which in turn implies $A(t) \to A(0)$. The boundedness condition on $A(t)$ implies that $\Theta^{(2)}_\mu$ remains bounded on $[0, T]$. To show that it remains constant, we study the derivatives $\partial_t \Theta^{(2)}_t$:
    \begin{align*}
        \partial_t \Theta^{(2)}_t(x, x') &= \nabla_{\theta_t} \Theta^{(2)}_t(x, x')^\top \partial_t \theta_t
    \end{align*}
    and we derive bounds for the norms of $\nabla_{\theta_t} \Theta^{(2)}_t(x, x')$ and $\partial_t \theta_t$:
    \begin{align}
        \left\lVert \partial_{W^{(1)}} \Theta^{(2)}_t(x, x') \right\rVert_2 &\leq \frac{2}{\sqrt{n_1}} c^2  \left \lVert \Theta^{(2)}_t \right \rVert_2 w^{(1)}(t) \label{eq:bound_dw1_theta} \\
        \left\lVert \partial_{W^{(0)}} \Theta^{(2)}_t(x, x') \right\rVert_2 &\leq \frac{2}{\sqrt{n_1}} c  \sup_{z \in \hat{\mathcal{N}}} \left\lVert z \right\rVert_2  a^{(1)}(t) \label{eq:bound_dw0_theta},
    \end{align}
    where $\hat{\mathcal{N}} = \bigcup_i \{x_i\} \cup \mathcal{N}(x_i)$. As $n_1 \to \infty$, we conclude that both \eqref{eq:bound_dw1_theta} and \eqref{eq:bound_dw0_theta} converge to $0$ because of the assumption $\left\lvert \hat{\mathcal{N}} \right\rvert < \infty$. Finally, since $A(t)$ is bounded, the same can be said of $\partial_t \theta$ as $n_1 \to \infty$. 
\end{proof}

\section{DenseDoorKey Environment}\label{sec:dense_doorkey}

\begin{figure}[htbp]
    \centering
    \begin{subfigure}[b]{0.45\columnwidth}
        \includegraphics[width=\textwidth]{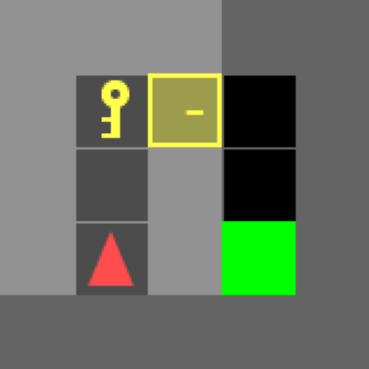}
        \caption{Initial grid potential $-0.01 \times l_1(p_0,g)$ \linebreak $ = -0.02$}
    \end{subfigure}
    \hfill
    \begin{subfigure}[b]{0.45\columnwidth}
        \includegraphics[width=\textwidth]{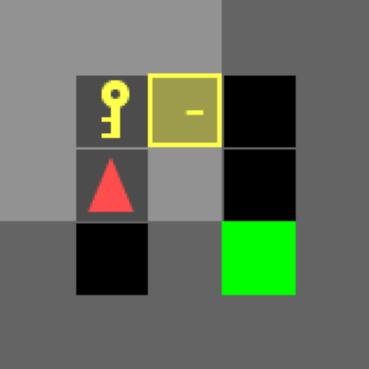}
        \caption{Step toward $g=\text{Key}$ $r_1 = 0.01 \times \Delta l_1(p_0,p_1,g)$ $=+0.01$}
    \end{subfigure}
    \begin{subfigure}[b]{0.45\columnwidth}
        \includegraphics[width=\textwidth]{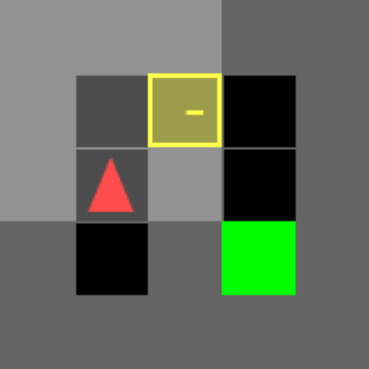}
        \caption{Subgoal 1: Pickup Key $r_{\text{subgoal}} = +0.2$}
    \end{subfigure}
    \hfill
    \begin{subfigure}[b]{0.45\columnwidth}
        \includegraphics[width=\textwidth]{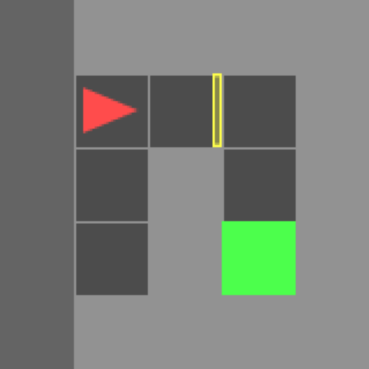}
        \caption{Subgoals 2: Unlock door $r_{\text{subgoal}} = +0.4$}
    \end{subfigure}
    \caption{Sequence of rewards in the \texttt{DenseDoorKey} environment. The agent is represented in red. The agent receives incremental potential-based distance rewards towards active targets, along with sparse intermediate subgoal rewards upon picking up the key ($+0.2$), and unlocking the door ($+0.4$).}
    \label{fig:densedoorkey_rewards}
\end{figure}

In the \texttt{DoorKey} gridworld environment, an agent must reach a target state located behind a locked door, which requires first finding and picking up a key. A non-zero reward $r_t > 0$ is obtained strictly upon reaching the final target state at terminal step $T$, whereas all intermediate transitions ($t < T$) yield $r_t = 0$. To overcome this reward sparsity, we introduce a new reward function that targets subgoals. This is a common trick in RL and it has been demonstrated to help with sample efficiency~\citep{okudo2021reward}. In other cases, intrinsic rewards can also alleviate the sparse reward problem~\citep{yuan2023airs}. The subgoals the we introduce are: 1) picking up the key, and 2) unlocking the door, to which we attribute a reward of $0.2$ and $0.4$ respectively. In addition, we include intermediate rewards based on the Manhattan distance ($l_1$): consider an agent in position $p_t$ who, after an action, transitions to position $p_{t+1}$; then, they obtain a reward equal to $r_{t+1}=0.01 \cdot \Delta l_1(p_t,p_{t+1},g)$ where $g$ is the position of the next immediate subgoal and $\Delta l_1(p_t,p_{t+1},g)=\left( l_1(p_t, g) - l_1(p_{t+1}, g) \right)$. Note that these intermediary rewards are positive when the agent moves closer to $g$ and negative otherwise. The reward for reaching the terminal goal (green cell) is $1$. We refer to the reader to Figure~\ref{fig:densedoorkey_rewards} for an illustration of the reward mechanism on a $5 \times 5$ grid.

\section{Mathematical model used by the solver}\label{sec:solver_model}
In reinforcement learning, active exploration traditionally relies on local, step-by-step environment rollouts or random trajectory sampling during training. In contrast, our mathematical optimization model \ref{eq:nearconvex_program} reformulates active exploration as a global search problem over the environment's state-action space. Our objective is to retrieve a high-value state $x = (s, a)$ by finding the corresponding grid and action. In this section, we detail the implementation of \ref{eq:nearconvex_program} within the RL training procedure. In this section, we describe in detail how \ref{eq:nearconvex_program} is implemented. Specifically, we detail for the \texttt{Minigrid} environment the constraints defining the set $\mathcal{C}$, and we also specify the explicit formulation of the constraints involving the kernel $\Theta_{\infty}^{(L)}$. To simplify notation, we let $[m] = \{1, \dots, m \}$, where $m \in \mathbb{N}$, and we let $[m]_k = \{k, \dots, m \}$, where $k \leq m, k \in \mathbb{N}$. In our \texttt{Minigrid} environment, the grid dimensions are $h \times w$, where $h$ denotes the height and $w$ denotes the width. 

Every \texttt{Minigrid} environment follows the convention that grids are tensors of shape $(h, w, 3)$. For any cell, the three values along the final dimension encode its object type, object colour, and state, respectively. For \texttt{DoorKey} grids, these values are mapped as follows:
\begin{align*}
    \text{object} &\rightarrow \{ \text{unseen}: 0, \text{empty}: 1, \text{wall}: 2, \text{door: 4}, \\
    &\quad\quad \text{key}: 5, \text{goal}: 8, \text{agent}: 10 \} \\
    \text{colour} &\rightarrow \{ \text{red}: 0, \text{green}: 1, \text{yellow}: 4, \text{grey}: 5 \} \\
    \text{state} &\rightarrow \{ \text{open}: 0, \text{closed}: 1, \text{locked}: 2 \}.
\end{align*}

Available actions follow the convention
\begin{align*}
    \text{turn left} &\rightarrow 0\\
    \text{turn right} &\rightarrow 1\\
    \text{move forward} &\rightarrow 2\\
    \text{pickup key} &\rightarrow 3\\
    \text{unlock door} &\rightarrow 5.
\end{align*}

\subsection{Grid constraints}
For each grid cell $(i,j) \in [h] \times [w]$ and $k \in [3]$, we introduce a  variable $s_{i,j,k} \in \mathbb{N}$ within
\begin{equation*}
    s_{i,j,k} \in 
    \begin{cases}
        [0, 10] &\text{if } k=1 \\
        [0, 5] &\text{if } k=2 \\
        [0, 2] &\text{if } k=3.
    \end{cases}
\end{equation*}
In simple words, these variables serve to describe the feature tuple $(s_{i,j,1}, s_{i,j,2}, s_{i,j,3})$ of cell $(i,j)$, where component $s_{i,j,1}$ denotes the object type, $s_{i,j,2}$ the object color, and $s_{i,j,3}$ the state..

We also introduce a variable $a \in \{0, 1, 2, 3, 5\}$ for the available actions. We add big-M constraints to determine the door and key positions. We also introduce binary variables $d_{i,j}, k_{i,j}$, for $(i,j) \in [h] \times [w]$, indicating the presence of a door or key in cell $(i,j)$, respectively. With these variables, we add constraints to ensure that there is one door and at most one key on the grid .
\begin{align*}
    s_{i,j,1} &\leq 4 + 10 \times (1 - d_{i,j} ) \tag{door in cell} \\
    s_{i,j,1} &\geq 4 - 10 \times (1 - d_{i,j} ) \\
    \sum_{(i,j)} d_{i,j} &= 1 \tag{exactly one door} \\
    s_{i,j,1} &\leq 5 + 10 \times (1 - k_{i,j} ) \tag{key in cell} \\
    s_{i,j,1} &\geq 5 - 10 \times (1 - k_{i,j} ) \\
    \sum_{(i,j)} k_{i,j} &\leq 1 \tag{at most one key}.
\end{align*}

We fix the cell at position $(h-1, w-1)$ to be a green goal square. In other words,
\begin{equation*}
    s_{h-1,w-1,1} = 8, \quad s_{h,w,2} = 1, \quad s_{h,w,3} = 0.
\end{equation*}

We introduce the binary variables $\text{agent}_{i,j} \in \{0, 1\}$ for $(i,j) \in [h-1]_2 \times [w-1]_2$ to determine the agent's position. In order to have only one agent on the grid, we add the constraint
\begin{equation*}
    \sum_{i=2}^{h-1} \sum_{j=2}^{w-1} \text{agent}_{i,j} = 1.
\end{equation*}

To determine if each cell is empty, we introduce the variables $\text{empty}_{i,j} \in \{0, 1\}$ for $(i,j) \in [h-1]_2 \times [w-1]_2$. We introduce the variables $\text{wall}_{i,j} \in \{0, 1\}$ for $(i,j) \in [h-1]_2 \times [w-2]_3$ to determine the presence of a wall. We note that \texttt{DoorKey} environments generate a separating wall containing a door at random between columns $j=3$ and $j=w-2$. We introduce the constraints
\begin{equation*}
    \text{agent}_{i,j} \leq 1 - \text{wall}_{i,j} \tag{agent not on wall}
\end{equation*}
to ensure the agent is not on a wall. The variables $\text{col}_j$ for $j \in [w-2]_3$ correspond to the presence of a separating wall in column $j$. The following constraints ensure 1) that there is one separating wall and 2) the separating wall consists only of wall cells or a door cell.
\begin{align*}
    \sum_{j \in [w]} \text{col}_j &= 1 \tag{1} \\
    \text{wall}_{i,j} + d_{i,j} &= \text{col}_{j} \tag{2} \\
\end{align*}

The agent orientation at cell $(i,j) \in [h-1]_2 \times [w-1]_2$ is determined through variables $\text{east}_{i,j} \in \{0,1\}$, $\text{south}_{i,j} \in \{0, 1\}$, $\text{west}_{i,j} \in \{0,1\}$, and $\text{north}_{i,j} \in \{0, 1\}$. We further require constraints for $(i,j) \in [h-1]_2 \times [w-2]_3$ to ensure that the agent is not on the key:
\begin{align*}
    \text{agent}_{i,j} &\leq 1 - k_{i,j}.
\end{align*}
The following constraint ensure the agent is not on the goal:
\begin{align*}
    \text{agent}_{h-1,w-1} &= 1 - \text{goal}_{h-1,w-1}.
\end{align*}
The agent orientation variables can only be active if there is an agent on the same cell:
\begin{equation*}
    \text{agent}_{i,j} = \text{east}_{i,j} + \text{south}_{i,j} + \text{west}_{i,j} + \text{north}_{i,j}.
\end{equation*}

We also fix the boundary cells of  the grid with $s_{i,j,1} = 2, s_{i,j,2} = 5, s_{i,j,3} = 0$. For each grid cell not on the boundary, we add two sets of constraints to determine its colour. The first set involves indices $(i,j) \in [h-1]_2 \times [w-2]_3$, corresponding to cells where the separating wall can lie:
\begin{align*}
    s_{i,j,1} &= 5 \times k_{i,j} + 2 \times \text{wall}_{i,j} \\
    &+ 4 \times d_{i,j} + 10 \times \text{agent}_{i,j} + \text{empty}_{i,j}.
\end{align*}

The second set of constraints involves indices $(i,j) \in [h-1]_2 \times \{2, w-1\}$, where the separating wall cannnot lie:
\begin{equation*}
    s_{i,j,1}  =  5 \times k_{i,j} + 10 \times \text{agent}_{i,j} + \text{empty}_{i,j}.
\end{equation*}

Finally, we introduce binary variables to model the door being open, closed, locked, and the key being carried. These variables are $\text{open}_{i,j} \in \{0,1\}$, $\text{closed}_{i,j} \in \{0, 1\}$, $\text{locked}_{i,j} \in \{0, 1\}$. For $(i,j) \in [h-1]_2, \times [w-2]_3$, we enforce the door to be in one of its three possible states if present in cell $(i,j)$:
\begin{equation*}
    d_{i,j} =  \text{open}_{i,j} + \text{closed}_{i,j} + \text{locked}_{i,j}.
\end{equation*}

We introduce a constraint that enforces the door to be locked if the agent does not carry the key, and a second constraint that forces to agent to be on the left-hand side of the door if it is not open; if the door is open, the agent can be on the right-hand side of the door. 

\begin{align*}
    \sum_{i=2}^{h-1} \sum_{j=3}^{w-2} \text{locked}_{i,j} &\geq \sum_{(i,j)} k_{i,j} \\
    \sum_{(k,l) \mid l \geq j} \text{agent}_{k,l} &\leq 1 - \text{col}_j + \sum_{(k,l)} \text{open}_{k,l} &j \in [w-2]_3.
\end{align*}
The second equation ensures that if the door is locked, the agent can only be found on the left-hand side of the separating wall. Otherwise, it is free to be on either side. The remaining grid constraints fix the state of the grid cells $(i,j) \in [h-1]_2 \times [w-2]_3$, i.e. a separating wall can be found: 
\begin{align*}
 s_{i,j,3} &= 1 \times \text{closed}_{i,j} + 2 \times \text{locked}_{i,j} \\
    &+ 3 \times \text{north}_{i,j} + 2 \times \text{west}_{i,j} + \text{south}_{i,j}  
\end{align*}
and of the cells $(i,j) \in [h-1]_2 \times \{2, w - 1\}$, i.e. outside the separating wall range of possible columns: 
\begin{align*}
    s_{i,j,3} &= 3 \times \text{north}_{i,j} + 2 \times \text{west}_{i,j} + \text{south}_{i,j} 
\end{align*}

All the variables and constraints defined in this section form the variable vector $x$ and the constraint set $\mathcal{C}$ appearing in \ref{eq:nearconvex_program}.

\subsection{Kernel constraints}\label{sec:kernel_constraints}
In this section, we introduce the remaining variables and constraints derived from \ref{eq:nearconvex_program}. The additional variables are 
\begin{align*}
    y_{i} &\in \mathbb{R}, & i &\in [n] \tag{embedding}\\
    \bar{e}_i &\in \mathbb{R}, &i &\in [n] \tag{normalization} \\
    \text{norm} &\in \mathbb{R}_{\geq 0} \\
    \text{asin}_{l,i} &\in \left[-\frac{\pi}{2}, \frac{\pi}{2} \right], &l &\in [L]_2, i \in [n] \\
    \rho_{l,i} &\in [-1, 1], &l &\in [L], i \in [n] \\
    \rho'_{l,i} &\in [-1, 1] &l &\in [L]_2, i \in [n] \\
    \Theta_{l, i} &\in [-1, 1], &l &\in [L], i \in [n],
\end{align*}
where variable $y$ corresponds to the variable $y = X x $ in \ref{eq:nearconvex_program}. We include constraint to model the data in $S^{n_0 - 1}$:
\begin{align*}
    x &= \text{concat} \left( \vect (s), a\right) \tag{$x = (s, a)$}\\
    \text{norm}^2 &=  \lVert x \rVert_2^2 \\
    y_i &= x_i^\top x &i &\in [n] \tag{dot product} \\
    \bar{e}_i \times \text{norm} &= \frac{y_i}{\lVert x_i \rVert_2} &i &\in [n]
    \tag{normalization}
\end{align*}
and constraints to model the kernel:
\begin{align*}
    \rho_{1, i} &= \bar{e}_i, &i &\in [n] \\
    \rho_{l, i} &= \text{asin}_{l, i} \times \frac{\sin(\text{asin}_{l,i})}{\pi} \\
    &+ \frac{\cos(\text{asin}_{l,i} )}{\pi} + \frac{\sin(\text{asin}_{l,i})}{2}, 
    &l &\in [L]_2, i \in [n]\\
    \rho'_{l,i} &= \frac{\text{asin}_{l,i} }{\pi} + \frac{1}{2}, &l &\in [L]_2, i \in [n] \\
    \sin(\text{asin}_{l,i}) &= \rho_{l-1, i}, &l &\in [L]_2, i \in [n] \\
    \Theta_{1, i} &= \rho_{1, i}, &i &\in [n] \\
    \Theta_{l, i} &= \frac{l-1}{l} \rho'_{l,i} \Theta_{l-1, i} + \frac{1}{l} \rho_{l, i}, &l &\in [L]_2, i \in [n].
\end{align*}

The variables $\Theta_{L, i}$ model the term $\Theta_{\infty}^{(L)}(y)$ in \ref{eq:nearconvex_program} through the equations
\begin{equation*}
    \Theta_{L ,i} = \Theta_{\infty}^{(L)}\left( \frac{x_i^\top x}{\lVert x_i \rVert_2 \lVert x \rVert_2} \right),
\end{equation*}
where $i \in [n]$. We remark that the matrix $\Theta_{\infty}^{(L)} \left( X X^\top \right)$ is known beforehand and is therefore a matrix of coefficients. In addition, the variables $\text{asin}_{l,i}$ model the $\arcsin$ terms in the definition of $\rho^{(l)}$. This is achieve through the constraint type $\sin(\alpha) = \beta$ for variables $\alpha$ and $\beta$, where the relationship $\arcsin(\beta) = \alpha$ is enforced. Moreover, functions $\rho^{(l)}(x)$ contain terms $\sqrt{1 - x^2}$, which can lead to numerical issues when computing their derivative with respect to $x$ near $-1$ or $1$. Instead, we model these terms with constraints $\cos(\text{asin}_{l.i})$, since $\sqrt{1 - \sin^2(\text{asin}_{l,i})} = \cos(\text{asin}_{l, i})$. The derivative of the $\cos$ function with respect to its input is bounded on its domain, thus numerically stable.

\section{Additional Experimental Details}

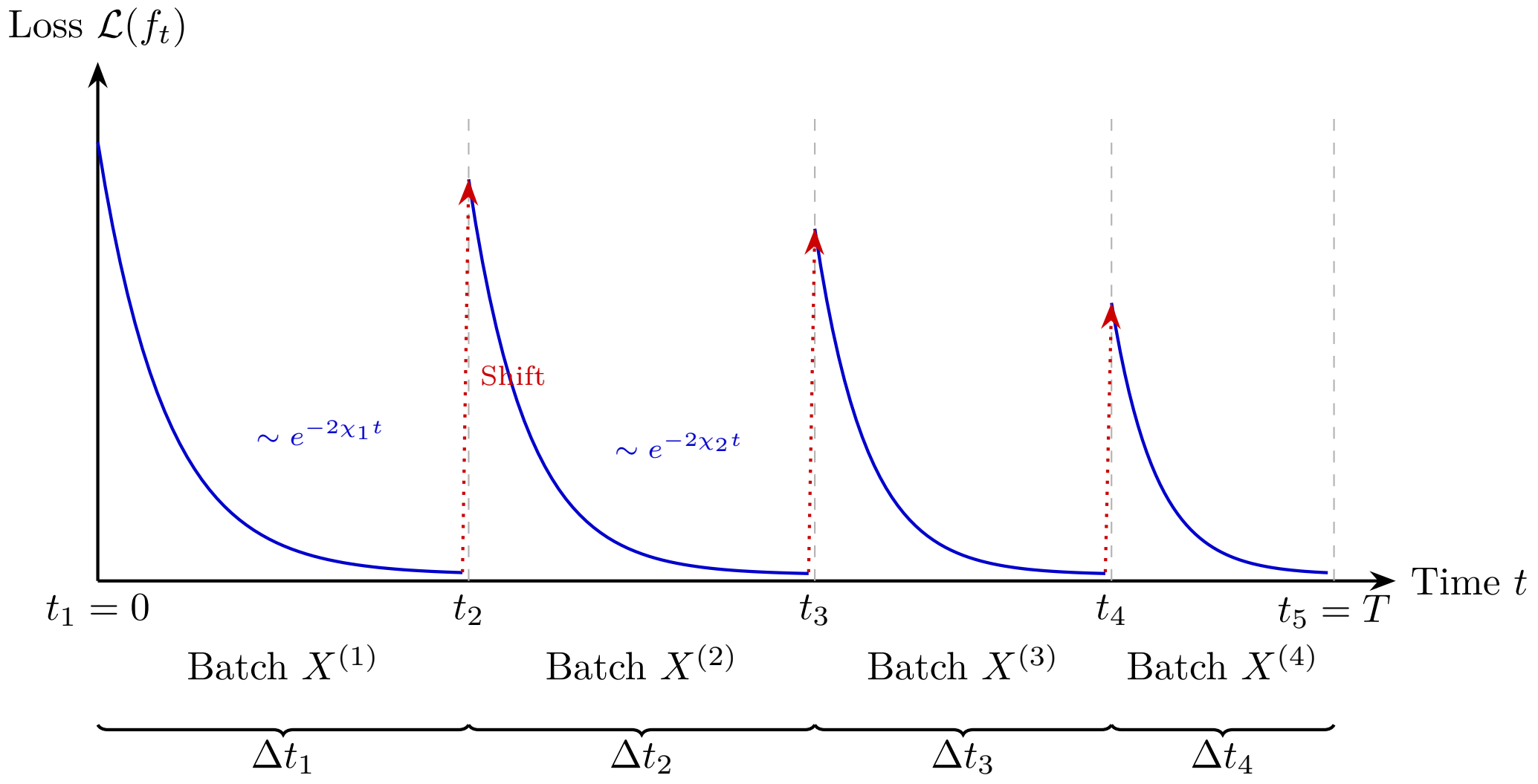
\begin{figure*}[htb]
\centering
\begin{tikzpicture}[
    scale=1.1,
    >=Stealth,
    batchline/.style={dashed, gray!60, thin},
    axisline/.style={->, thick},
    losscurve/.style={thick, blue!80!black}
]
    \draw[axisline] (0,0) -- (10.5,0) node[right] {Time $t$};
    \draw[axisline] (0,0) -- (0,4.2) node[above] {Loss $\mathcal{L}(f_t)$};

    \coordinate (T1) at (0,0);
    \coordinate (T2) at (3.0,0);
    \coordinate (T3) at (5.8,0);
    \coordinate (T4) at (8.2,0);
    \coordinate (T5) at (10.0,0);

    \draw[batchline] (3.0,0) -- (3.0,3.8);
    \draw[batchline] (5.8,0) -- (5.8,3.8);
    \draw[batchline] (8.2,0) -- (8.2,3.8);
    \draw[batchline] (10.0,0) -- (10.0,3.8);

    \node[below] at (T1) {$t_1=0$};
    \node[below] at (T2) {$t_2$};
    \node[below] at (T3) {$t_3$};
    \node[below] at (T4) {$t_4$};
    \node[below] at (T5) {$t_5=T$};

    \node[anchor=north] at (1.5,-0.4) {Batch $X^{(1)}$};
    \node[anchor=north] at (4.4,-0.4) {Batch $X^{(2)}$};
    \node[anchor=north] at (7.0,-0.4) {Batch $X^{(3)}$};
    \node[anchor=north] at (9.1,-0.4) {Batch $X^{(4)}$};

    \draw[decorate, decoration={brace, mirror, raise=2pt}, thick] (0,-1.1) -- (3.0,-1.1) node[midway,below=2pt] {$\Delta t_1$};
    \draw[decorate, decoration={brace, mirror, raise=2pt}, thick] (3.0,-1.1) -- (5.8,-1.1) node[midway,below=2pt] {$\Delta t_2$};
    \draw[decorate, decoration={brace, mirror, raise=2pt}, thick] (5.8,-1.1) -- (8.2,-1.1) node[midway,below=2pt] {$\Delta t_3$};
    \draw[decorate, decoration={brace, mirror, raise=2pt}, thick] (8.2,-1.1) -- (10.0,-1.1) node[midway,below=2pt] {$\Delta t_4$};

    \draw[losscurve] plot[domain=0:2.95, samples=50] (\x, {3.5*exp(-1.8*\x) + 0.05});
    \draw[losscurve] plot[domain=3.0:5.75, samples=50] (\x, {3.2*exp(-2.1*(\x-3.0)) + 0.05});
    \draw[losscurve] plot[domain=5.8:8.15, samples=50] (\x, {2.8*exp(-2.4*(\x-5.8)) + 0.05});
    \draw[losscurve] plot[domain=8.2:9.95, samples=50] (\x, {2.2*exp(-2.8*(\x-8.2)) + 0.05});

    \draw[dotted, red!80!black, thick, ->] (2.95, 0.07) -- (3.0, 3.25) node[midway, right] {\scriptsize Shift};
    \draw[dotted, red!80!black, thick, ->] (5.75, 0.07) -- (5.8, 2.85);
    \draw[dotted, red!80!black, thick, ->] (8.15, 0.07) -- (8.2, 2.25);

    \node[blue!80!black] at (1.8, 1.2) {\scriptsize $\sim e^{-2\chi_1 t}$};
    \node[blue!80!black] at (4.7, 1.1) {\scriptsize $\sim e^{-2\chi_2 t}$};

\end{tikzpicture}
\caption{Piecewise continuous NTK gradient flow across mini-batch transitions. At each stopping time $t_b$, mini-batch replacement causes a loss spike due to distribution shift, followed by exponential decay governed by the local strong convexity constant $\chi_b = \lambda_{\min}\left(\Theta_\infty^{(L)}\left(X^{(b)}X^{(b)\top}\right)\right) > 0$.}
\label{fig:tikz_minibatch_flow}
\end{figure*}

In our experiments, we train on a sequence of mini-batches $\{X^{(b)}\}_{b=1}^B$. The default approach in RL is to train on data selected from the replay buffer uniformly at random, or through some prioritized replay mechanism. In these instances, the replay buffer data goes through a distribution shift. In our experiments, we noticed that by freezing exploration following the introduction of new data, we obtained a more stable training when we introduced decreasing time increments. This idea was motivated by the literature on stochastic gradient descent (SGD), where, given a fixed dataset, a schedule of decreasing learning rates guarantees convergence to a global minimum\footnote{With respect to the full dataset.} in the $n$-smooth and $\chi$-strongly convex setting~\citep{garrigos2023handbook}. Because our overparameterized networks satisfy these convergence assumptions, we could expect this behaviour on a fixed dataset if batches are selected uniformly at random. However, distribution shift breaks the uniformity assumption by giving more weight to tuples obtained via an older policy or a prioritized sampling. For this reason, we periodically collect new samples and between each data collection phase, we perform batch training through the closed-form solution to the TDF (Theorem~\ref{thm:td_closed_formulation}). The training procedure is illustrated in Figure~\ref{fig:tikz_minibatch_flow}.

\begin{figure*}[htb]
\centering
\begin{tikzpicture}[
    >=Stealth,
    node distance=1.5cm and 5.0cm,
    block/.style={rectangle, draw=black!80, thick, fill=blue!5, rounded corners=3pt, align=center, inner sep=8pt, minimum width=3.2cm, minimum height=1.2cm},
    solverblock/.style={rectangle, draw=red!80!black, thick, fill=red!5, rounded corners=3pt, align=center, inner sep=8pt, minimum width=3.4cm, minimum height=1.2cm},
    proxyblock/.style={rectangle, draw=green!60!black, thick, fill=green!5, rounded corners=3pt, align=center, inner sep=8pt, minimum width=3.4cm, minimum height=1.2cm},
    line/.style={draw, thick, ->},
    labeltext/.style={font=\small, align=center}
]

    \node [block] (env) {\textbf{Environment}\\ \texttt{DenseDoorKey} ($8\times 8$)};
    \node [block, right=of env] (buffer) {\textbf{Replay Buffer $\mathcal{B}$}\\ Unique $(s_i, a_i)$ Tuples};
    \node [proxyblock, below=of buffer] (ntk) {\textbf{Analytical NTK Proxy}\\ Closed-form $Q_\infty(s,a)$\\ Strongly Convex \& Smooth};
    \node [solverblock, below=of env] (solver) {\textbf{MINLP Solver (Gurobi)}\\ Mathematical Program $(P_\omega^\pm)$\\ + No-Good / Hamming Cuts};

    \draw [line] (env) -- node[above, labeltext] {Transitions\\$(s, a, r, s')$} (buffer);
    \draw [line] (buffer) -- node[right, labeltext] {Mini-Batch $X^{(b)}$\\Data Distribution} (ntk);
    \draw [line] (ntk) -- node[below, labeltext] {Surrogate Loss Surface\\$\mathcal{L}_{\text{TD}}(s,a)$} (solver);
    \draw [line] (buffer) -- node[left, labeltext] {Excluded Set $\mathcal{B}$\\Diversity Constraints} (solver);
    \draw [line] (solver) -- node[left, labeltext] {Target Candidate $(s^*, a^*)$\\Active Exploration Step} (env);

    \begin{scope}[on background layer]
        \node [draw=black!40, dashed, thick, fill=gray!5, fit=(ntk) (solver), inner sep=12pt, label=below:\small{\textbf{Solver-Augmented Exploration Callback Loop}}] (loopbox) {};
    \end{scope}

\end{tikzpicture}
\caption{Architecture of the solver-augmented exploration callback loop. The analytical NTK proxy constructs a smooth, strongly convex value landscape from buffer mini-batches. The MINLP solver optimizes over this landscape subject to No-Good cuts, identifying high-error state-action targets $(s^*, a^*)$ for environment rollouts.}
\label{fig:tikz_solver_architecture}
\end{figure*}
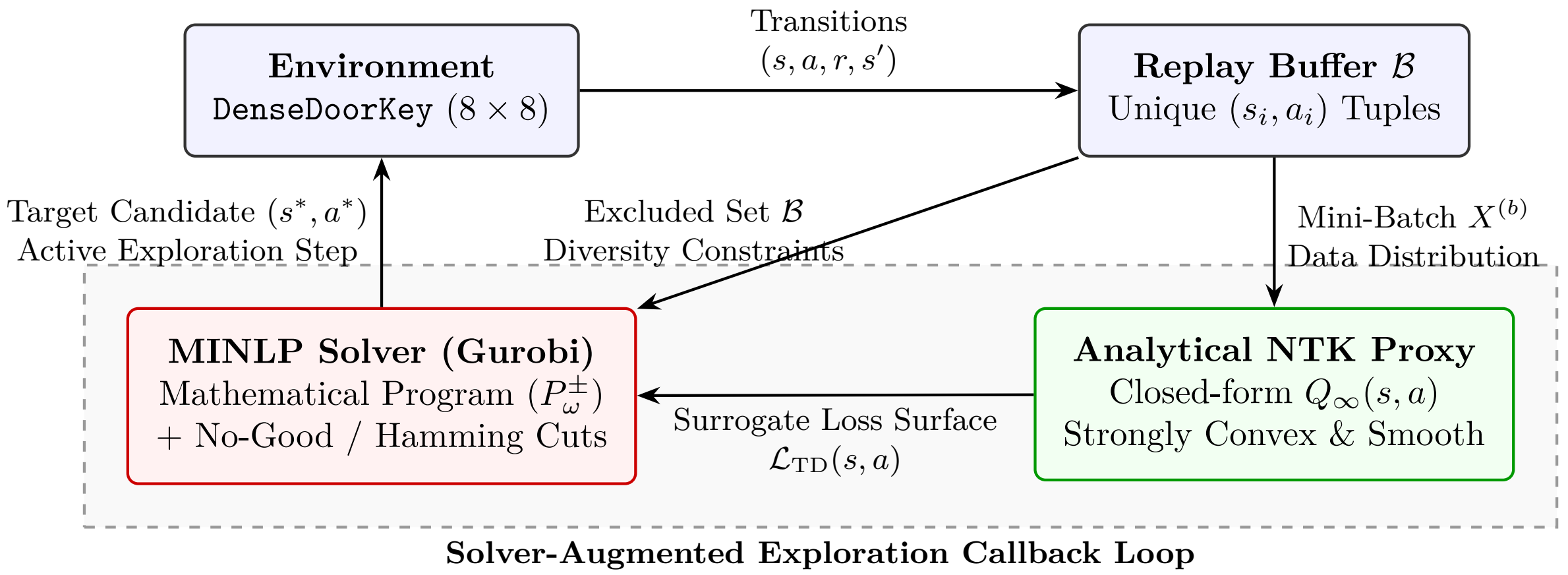

In addition to the optimization model described in Appendix~\ref{sec:solver_model}, we summarize the callback loop used during the solver-augmented exploration in Figure~\ref{fig:tikz_solver_architecture}. While we made minimal use of diversity constraints (i.e. no-good cuts and hamming cuts), the solver-enhanced approach allows for further exploration of diversity-related ideas that are not available without the closed-form solution of the NTK. We compute the average running time of the solver callback procedure, i.e. the time to solve the optimization problem. The average time was $0.097$ seconds with a standard deviation of $0.028$ seconds. 

\section{Necessity of KKT conditions}
In this section, we show that the KKT conditions for \ref{eq:nearconvex_program} are necessary for a local optimum with respect to the primal variables (e.g., $x$ and $y$). We begin by providing the constraint qualification required in the proof of necessity~\citep{andreani2012relaxed}. 

\begin{deff}[RCPLD]\label{def:cpld}
    Denote the equality constraints with $h_i(x) = 0, i \in \mathcal{I}$ and the inequality constraints $g_j(x) \leq 0, j \in \mathcal{J}$, and let
    $\mathcal{J}(x^\star) = \{ j \in \mathcal{J} : g_j(x^\star) = 0 \}$ be the set of
    constraints active at $x^\star$. The relaxed constant positive linear dependence (RCPLD) condition holds if there exists a neighbourhood $B(x^\star, \delta) = \{ x : \lVert x - x^\star \rVert_2 < \delta \}$ for some $\delta > 0$ such that 1) the set $\{ \nabla h_i(x) | i \in \mathcal{I} \}$ has constant rank for $x \in B(x^\star, \delta)$, and 2) for every basis of $\{ \nabla h_i(x^\star) \mid \mathcal{I} \}$ with index set $I \subseteq \mathcal{I}$ and for any subset $J \subseteq \mathcal{J}(x^\star)$, 
    the following implication is satisfied:
     \begin{align*}
        &\exists\, (\lambda, \mu) \neq 0,\ \mu \geq 0 \ :\
        \sum_{i \in I} \lambda_i \nabla h_i(x^\star)
        + \sum_{j \in J} \mu_j \nabla g_j(x^\star) = 0 \\
        \implies\quad
        &\forall x \in B(x^\star, \delta) \; \exists\, (\alpha, \beta) \neq 0 \ :\
        \sum_{i \in I} \alpha_i \nabla h_i(x)
        + \sum_{j \in J} \beta_j \nabla g_j(x) = 0.
    \end{align*}
\end{deff}

Definition~\ref{def:cpld}~\citep{andreani2005relation} is a relaxed version of the CPLD constraint qualification introduced by \citet{andreani2005relation}. With Definition~\ref{def:cpld}, we state the circumstances under which the KKT conditions are necessary for a local optimum in Theorem~\ref{thm:kkt_necessary}.

\begin{thm}\label{thm:kkt_necessary}
    For a primal variable $v$ of \ref{eq:nearconvex_program}, let $v^\star$denote the corresponding variable value in a (feasible) local optimum. Suppose the following:
    \begin{enumerate}
        \item $x^\star$ is such that $\lVert x \rVert_2 > 0$ and $x^\star \neq \pm x_i$ for all $i \in [n]$;
        \item $x_i^\top x_j \in [-1, 1]$ for $i,j \in [n]$ and that $x_i^\top x_j = 1$ only if $i=j$;
        \item $L \geq 2$ so that $\Theta_{\infty}^{(L)} \left( X X^\top \right)$ is invertible. 
    \end{enumerate}
    Then, the local optimum satisfies the KKT conditions for some dual coefficients multiplying the equality and inequality constraints. 
\end{thm}
\begin{proof}
    From the stationarity condition, we get that a linear combination of the gradients of the equality constraints and the active inequality constraints is equal to $0$ only if the dual coefficients are all equal to $0$, except (possibly) the dual coefficients of $\mathcal{C}(s, a)$. However, $\mathcal{C}(s, a)$ corresponds to linear equalities and inequalities. A linear combination of an arbitrary subset of these gradients that is equal to $0$ at $x^\star$ leads to a linear combination that is equal to $0$ at $x$. Because of this fact, the rank of $\{\nabla h_i(x) \mid i \in \mathcal{I} \}$ is constant for $x \in B(x^\star, \delta)$. We conclude that the constraints of \eqref{eq:nearconvex_program} satisfy RCPLD and that the KKT conditions are necessary for the local optimal solution. 
\end{proof}

In the statement of Theorem~\ref{thm:kkt_necessary}, the vectors $x_i,~i \in [n],$ are normalized with unit norm, but since the unnormalized version $\Theta_{\infty}^{(L)}$ is homogeneous component-wise~\citep{jacot2018neural}, the statement holds in the general sense by assuming the $x_i$'s and $x$ are not pairwise colinear. The proof of Theorem~\ref{thm:kkt_necessary} also holds in the more general case where the constraints of $\mathcal{C}(s, a)$ are not linear; in this instance, one needs some regularity condition like quasi-normality constraint qualification. Theorem~\ref{thm:kkt_necessary} guarantees that in a MIP approach to solving \ref{eq:nearconvex_program} (e.g. a spatial branch-and-bound), the upper bound\footnote{Assuming minimization in \ref{eq:nearconvex_program}.} search in the root node is well-behaved when using first-order approximations. For example, primal-dual interior point solvers like IPOPT will converge to stationary points (local minimum or maximum, saddle point) that are feasible, and any local optimum will satisfy the KKT conditions. As shown in Corollary~\ref{cor:kkt_necessary}, the first-order necessary condition extends to any local node in a MIP approach whenever the depth $L$ and the size of the dataset $n$ are large enough compared to the number of variables required to model the state space.

\begin{cor}\label{cor:kkt_necessary}
    Under the same assumptions as Theorem~\ref{thm:kkt_necessary}, and if we propagate the bounds of variable $\text{asin}_{l,i}, \rho_{L,i}, \rho'_{l,i}, \Theta_{l,i}$, a local node will also satisfy RCPLD whenever
    \begin{equation*}
        (4 L - 2) n + 1 > |x| - m,
    \end{equation*}
    where $|x| = |s| + 1$ is the length of variable $x$ and $m$ is the number of equality constraints in $\mathcal{C}(s, a)$.
\end{cor}

\begin{proof}
    The gradients $\nabla g_j(x^\star)$ for $j \in J$ induce a cone $C_J$, while the gradients $\nabla h_i(x^\star)$ for $i \in \mathcal{I}$ induce a subspace $H_{\mathcal{I}}(x^\star)$. We suppose $C_J \cap H_{\mathcal{I}}(x^\star) \neq \{0\}$ and we observe that
    \begin{align}\label{eq:span_intersection}
        \Span( \{ \nabla g_j(y) \mid j \in J \} ) \cap H_{\mathcal{I}}(y) &= \Span( \{ \nabla g_j(x^\star) \mid j \in J \} ) \cap H_{\mathcal{I}}(y).
    \end{align}
    For vector subspaces $V$ and $W$, 
    \begin{equation*}
        \dim(V \cap W) = \dim(V) + \dim(W) - \dim(V + W).
    \end{equation*}
    Therefore, subspace~\eqref{eq:span_intersection} has dimension strictly greater than zero if
    \begin{align*}
        &\phantom{} & \dim( H_\mathcal{I}(y)) + \dim( \Span(\{ \nabla g_j(y) \mid j \in J \} ) ) &> |y| \\
        &\iff &  4L n + |y_x| + 1 + m + (4L-2)n &> 4Ln + 1 + 2 |y_x| \\
        &\iff & (8L-2)n + |y_x| + 1 + m &> 4Ln + 1 + 2 | y_x| \\
        &\iff & (4L-2) n + 1 &> |y_x| - m,
    \end{align*}
    where $y_x$ is the section of $y$ indexed by variable $x$. Given our initial assumption, RCPLD is satisfied in a neighbourhood of $x^\star$.
\end{proof}

\end{document}